\documentclass[journal]{IEEEtran}

\usepackage{cite}
\usepackage{graphicx}
\usepackage{subfigure}
\usepackage{amsmath}
\usepackage{amssymb}
\usepackage{tabularx}
\usepackage[table,xcdraw]{xcolor}
\usepackage{multirow}
\usepackage{color}
\usepackage{algorithm}
\usepackage{booktabs}
\usepackage{hhline}
\usepackage{lineno}
\usepackage{colortbl}
\usepackage{amsthm}
\usepackage[colorlinks]{hyperref}
\usepackage[]{algpseudocode}

\newcommand{\RN}[1]{\textup{\uppercase\expandafter{\romannumeral#1}}}

\definecolor{dbcolor}{rgb}{0,0,1}

\makeatletter
\def\hlinewd#1{%
\noalign{\ifnum0=`}\fi\hrule \@height #1 \futurelet
\reserved@a\@xhline}
\begin{document}
\title{DIML/CVL RGB-D Dataset: 2M RGB-D Images of Natural Indoor and Outdoor Scenes}
\author{Jaehoon Cho,~\IEEEmembership{Student Member,~IEEE,}
	Dongbo Min,~\IEEEmembership{Senior Member,~IEEE,}\\
	Youngjung Kim,~\IEEEmembership{Member,~IEEE,}
	and Kwanghoon Sohn,~\IEEEmembership{Senior Member,~IEEE}\\
	\url{https://dimlrgbd.github.io/}}
\markboth{}{}
\maketitle
\IEEEpeerreviewmaketitle

\section{Overview}
This manual is intended to provide a detailed description about the DIML/CVL RGB-D dataset.
This dataset is comprised of 2M color images and their corresponding depth maps from a great variety of natural indoor and outdoor scenes.
The indoor dataset was constructed using the Microsoft Kinect v2~\cite{kinect}, while the outdoor dataset was built using the stereo cameras (ZED stereo camera~\cite{zed} and built-in stereo camera).
Table~\ref{tab:3} summarizes the details of our dataset, including acquisition, processing, format, and toolbox. Refer to Section II and III for more details.
The detailed camera specifications and RGB-D sample data are shown in Table~\ref{tab:1} and~\ref{tab:2}, respectively.

This dataset was taken from fall 2015 to summer 2017, in South Korea.
We built it for the purpose of promoting depth-related applications based on deep learning.
Our dataset differs from existing ones in the following aspects:
\begin{enumerate}
    \item It is comprised of both indoor and outdoor RGB-D data.
    \item Large-scale RGB-D dataset is provided, consisting of 1M indoor data and 1M outdoor data.
    \item Unlike existing outdoor data using stereo cameras mounted on a moving vehicle, ours was taken using hand-held stereo cameras.
    \item Confidence maps are provided together to quantify the accuracy of depth maps computed from stereo camera in outdoor scenes.
\end{enumerate}

\emph{Remark}: We are now in the process of analyzing the characteristics of the DIML/CVL RGB-D dataset.
For a timely impact on research communities, we first release a subset of the dataset before completing the analysis. The comprehensive report including performance analysis will be released later with the complete 2M RGB-D data.

\begin{table*}
    \centering
    \caption{Summary of DIML/CVL RGB-D Dataset.}
    \vspace{-3pt}
    \label{my-label}
    \begin{tabular}{l|l|l}
        \toprule
        & {\bf Indoor dataset}      & {\bf Outdoor dataset}  \\ \hline
        \midrule
        Data acquisition    &  Microsoft Kinect v2      & Stereo camera (ZED and built-in camera)        \\
        \midrule
        Data processing     & Calibration using the Kinect SDK~\cite{kinect}    & Calibration and rectification using Caltech toolbox~\cite{caltech} \\
        & Calibration using `iai$\_$kinect v2 libary~\cite{iai}                             & Stereo matching~\cite{Zbontar2015}  \\
        & Color guided depth upsampling~\cite{Kim2016}                                                           & Confidence estimation~\cite{Kim2018}  \\
        \midrule
        Data format         & \bf{Color images}                     & \bf{Color images}   \\
                            & - Cropped images                      & - Rectified left and right images \\
                            & \bf{Depth maps}                       & \bf{Disparity, depth, and confidence map} \\
                            & - Original depth map                  & - Left disparity and depth map \\
                            & - Projected depth map                 & - Left Confidence map \\
                            & - Hole-filled depth map               & \bf{Calibration parameters} \\
                            & \bf{Calibration parameters}           & - Intrinsic/extrinsic parameters for stereo camera \\
                            & - Intrinsic/extrinsic parameter for color and IR cameras               &                          \\
                            & - Shift parameter between IR and depth map                             &                                                    \\
        \midrule
       {MATLAB Toolbox }
                            & Fill depth                   &  Confidence estimation                   \\

        \bottomrule
    \end{tabular}\label{tab:3}
\end{table*}

\begin{table*}
    \centering
    \caption{Camera specifications used for data acquisition.}
        \vspace{-3pt}
    \resizebox{0.95\textwidth}{!}{
        \begin{tabular}{ccccc}
            \toprule
            {\bf } & {\bf Kinect v2~\cite{kinect}} & {\bf  ZED stereo~\cite{zed}}  & {\bf Built-in stereo} \\ \hline
            \midrule
            & \raisebox{-\totalheight}{\includegraphics[width=0.2\textwidth]{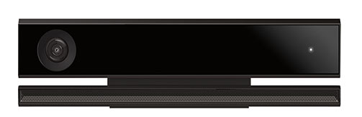}} & \raisebox{-\totalheight}{\includegraphics[width=0.2\textwidth]{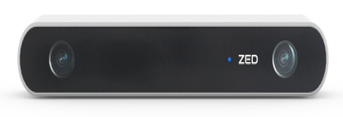}} & \raisebox{-\totalheight}{\includegraphics[width=0.2\textwidth]{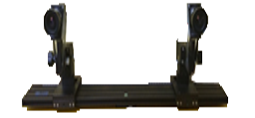}}   \\
            \midrule

            \multirow{2}{*}{Color Resolution}    & \multirow{2}{*}{1920 x 1080} & 1920 x 1080 & 1920 x 1080              \\
            &                                    & 1280 x 720  & 1280 x 720              \\
            \midrule
            \multirow{2}{*}{Depth Resolution}    & \multirow{2}{*}{512 x 424} & 1920 x 1080 & 1920 x 1080              \\
            &                               & 1280 x 720  & 1280 x 720              \\
            \midrule
            Range of Depth                       & 0.5 - 8.0 m                  & 0.5 - 20 m  & 2 - 80 m              \\
            \midrule
            Baseline                             & -                            & 12 cm       & 40 cm              \\
            \midrule
            Focal length                         &                              & 2.8 mm      & 3.5 mm              \\
            \midrule
            \multirow{2}{*}{Data acquisition}    & \multirow{2}{*}{Color and IR images, depth map}          & Left and right color images & Left and right color images              \\
            &                                           & Depth and confidence maps & Depth and confidence maps              \\
            \bottomrule
        \end{tabular}}\label{tab:1}
\end{table*}

    \vspace{-3pt}
    
\begin{table*}
    \centering
    \caption{Sample data from the DIML/CVL RGB-D dataset}
        \resizebox{0.98\textwidth}{!}{
    \begin{tabular}{cccc|cccc}
        \toprule
        \multicolumn{4}{c}{\bf Indoor}  &\multicolumn{4}{c}{\bf Outdoor}    \\  \cline{1-8}

        {\bf Color } & {\bf Depth} & {\bf Color } & {\bf Depth} & {\bf Left } & {\bf Right} & {\bf Disparity } & {\bf Confidence}\\ \hline

        \midrule

        \raisebox{-\totalheight}{\includegraphics[width=0.10\textwidth]{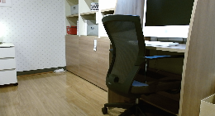}} &          \raisebox{-\totalheight}{\includegraphics[width=0.10\textwidth]{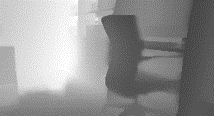}} & \raisebox{-\totalheight}{\includegraphics[width=0.10\textwidth]{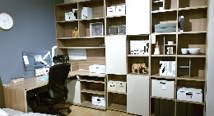}} &       \raisebox{-\totalheight}{\includegraphics[width=0.10\textwidth]{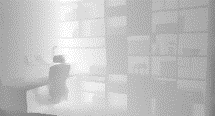}} &

        \raisebox{-\totalheight}{\includegraphics[width=0.10\textwidth]{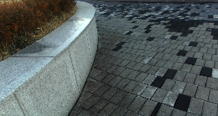}} &         \raisebox{-\totalheight}{\includegraphics[width=0.10\textwidth]{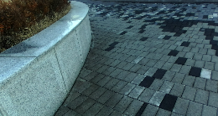}} & \raisebox{-\totalheight}{\includegraphics[width=0.10\textwidth]{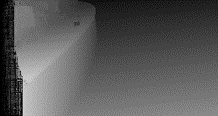}} &         \raisebox{-\totalheight}{\includegraphics[width=0.10\textwidth]{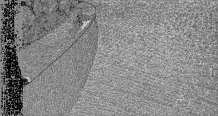}} \\

        \raisebox{-\totalheight}{\includegraphics[width=0.10\textwidth]{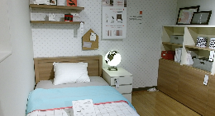}} &          \raisebox{-\totalheight}{\includegraphics[width=0.10\textwidth]{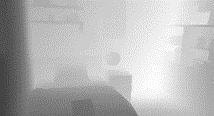}} & \raisebox{-\totalheight}{\includegraphics[width=0.10\textwidth]{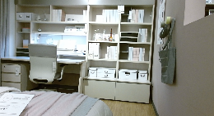}} &       \raisebox{-\totalheight}{\includegraphics[width=0.10\textwidth]{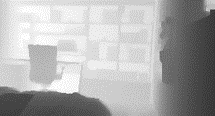}} &

        \raisebox{-\totalheight}{\includegraphics[width=0.10\textwidth]{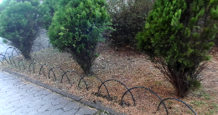}} &         \raisebox{-\totalheight}{\includegraphics[width=0.10\textwidth]{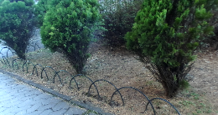}} & \raisebox{-\totalheight}{\includegraphics[width=0.10\textwidth]{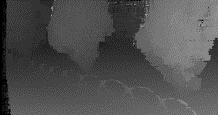}} &         \raisebox{-\totalheight}{\includegraphics[width=0.10\textwidth]{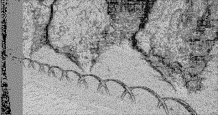}} \\

        \raisebox{-\totalheight}{\includegraphics[width=0.10\textwidth]{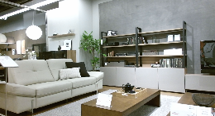}} &          \raisebox{-\totalheight}{\includegraphics[width=0.10\textwidth]{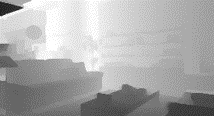}} & \raisebox{-\totalheight}{\includegraphics[width=0.10\textwidth]{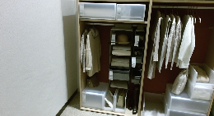}} &       \raisebox{-\totalheight}{\includegraphics[width=0.10\textwidth]{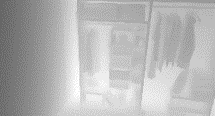}} &

        \raisebox{-\totalheight}{\includegraphics[width=0.10\textwidth]{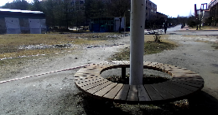}} &         \raisebox{-\totalheight}{\includegraphics[width=0.10\textwidth]{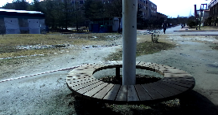}} & \raisebox{-\totalheight}{\includegraphics[width=0.10\textwidth]{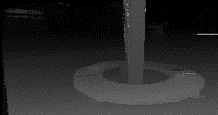}} &         \raisebox{-\totalheight}{\includegraphics[width=0.10\textwidth]{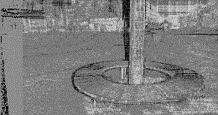}} \\

        \raisebox{-\totalheight}{\includegraphics[width=0.10\textwidth]{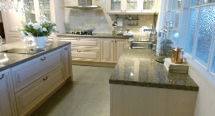}} &          \raisebox{-\totalheight}{\includegraphics[width=0.10\textwidth]{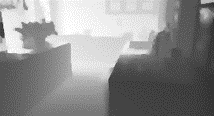}} & \raisebox{-\totalheight}{\includegraphics[width=0.10\textwidth]{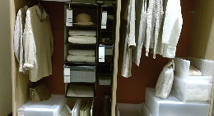}} &       \raisebox{-\totalheight}{\includegraphics[width=0.10\textwidth]{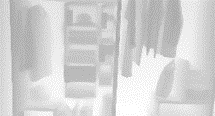}} &

        \raisebox{-\totalheight}{\includegraphics[width=0.10\textwidth]{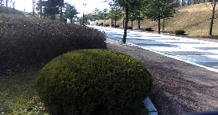}} &         \raisebox{-\totalheight}{\includegraphics[width=0.10\textwidth]{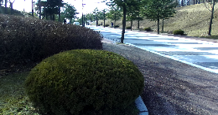}} & \raisebox{-\totalheight}{\includegraphics[width=0.10\textwidth]{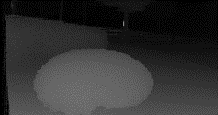}} &         \raisebox{-\totalheight}{\includegraphics[width=0.10\textwidth]{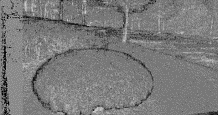}} \\

        \raisebox{-\totalheight}{\includegraphics[width=0.10\textwidth]{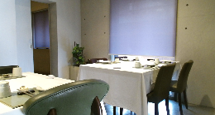}} &          \raisebox{-\totalheight}{\includegraphics[width=0.10\textwidth]{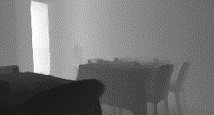}} & \raisebox{-\totalheight}{\includegraphics[width=0.10\textwidth]{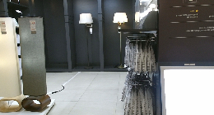}} &          \raisebox{-\totalheight}{\includegraphics[width=0.10\textwidth]{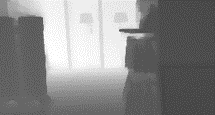}} &

        \raisebox{-\totalheight}{\includegraphics[width=0.10\textwidth]{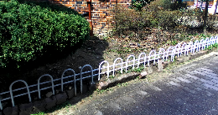}} &         \raisebox{-\totalheight}{\includegraphics[width=0.10\textwidth]{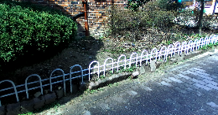}} & \raisebox{-\totalheight}{\includegraphics[width=0.10\textwidth]{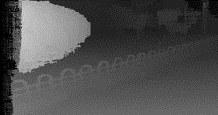}} &         \raisebox{-\totalheight}{\includegraphics[width=0.10\textwidth]{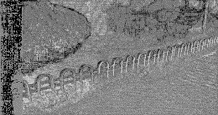}} \\

        \raisebox{-\totalheight}{\includegraphics[width=0.10\textwidth]{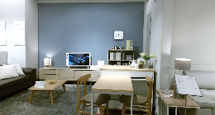}} &         \raisebox{-\totalheight}{\includegraphics[width=0.10\textwidth]{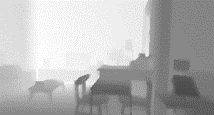}} & \raisebox{-\totalheight}{\includegraphics[width=0.10\textwidth]{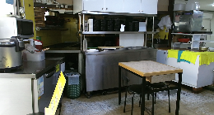}} &         \raisebox{-\totalheight}{\includegraphics[width=0.10\textwidth]{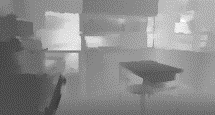}} &

        \raisebox{-\totalheight}{\includegraphics[width=0.10\textwidth]{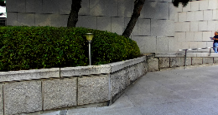}} &         \raisebox{-\totalheight}{\includegraphics[width=0.10\textwidth]{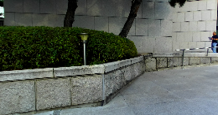}} & \raisebox{-\totalheight}{\includegraphics[width=0.10\textwidth]{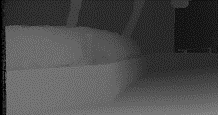}} &         \raisebox{-\totalheight}{\includegraphics[width=0.10\textwidth]{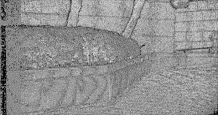}} \\

        \raisebox{-\totalheight}{\includegraphics[width=0.10\textwidth]{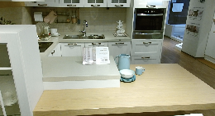}} &         \raisebox{-\totalheight}{\includegraphics[width=0.10\textwidth]{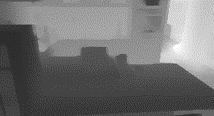}} & \raisebox{-\totalheight}{\includegraphics[width=0.10\textwidth]{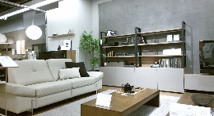}} &         \raisebox{-\totalheight}{\includegraphics[width=0.10\textwidth]{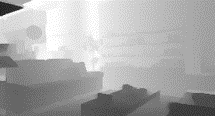}} &

        \raisebox{-\totalheight}{\includegraphics[width=0.10\textwidth]{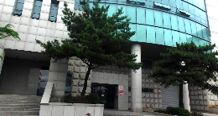}} &         \raisebox{-\totalheight}{\includegraphics[width=0.10\textwidth]{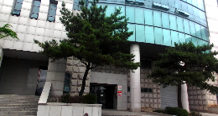}} & \raisebox{-\totalheight}{\includegraphics[width=0.10\textwidth]{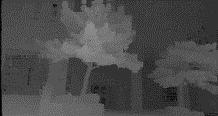}} &         \raisebox{-\totalheight}{\includegraphics[width=0.10\textwidth]{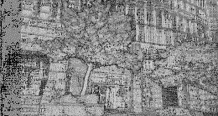}} \\

        \raisebox{-\totalheight}{\includegraphics[width=0.10\textwidth]{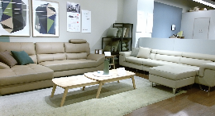}} &         \raisebox{-\totalheight}{\includegraphics[width=0.10\textwidth]{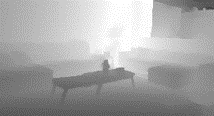}} & \raisebox{-\totalheight}{\includegraphics[width=0.10\textwidth]{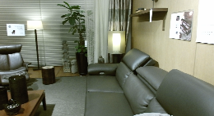}} &         \raisebox{-\totalheight}{\includegraphics[width=0.10\textwidth]{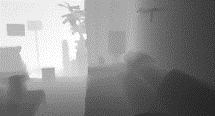}} &

        \raisebox{-\totalheight}{\includegraphics[width=0.10\textwidth]{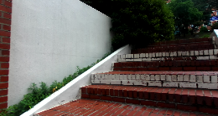}} &         \raisebox{-\totalheight}{\includegraphics[width=0.10\textwidth]{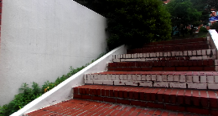}} & \raisebox{-\totalheight}{\includegraphics[width=0.10\textwidth]{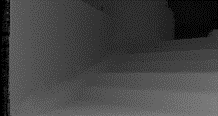}} &         \raisebox{-\totalheight}{\includegraphics[width=0.10\textwidth]{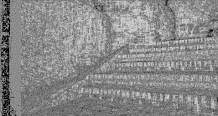}}
        \\
        \hline
        \bottomrule
    \end{tabular}}\label{tab:2}
\end{table*}
    \vspace{-3pt}

\section{Indoor RGB-D dataset}
\subsection{Data Acquisition}

\subsubsection{Hardware setup}
\begin{figure}
    \centering
    \renewcommand{\thesubfigure}{}
    \subfigure[]
    {\includegraphics[width=0.31\linewidth]{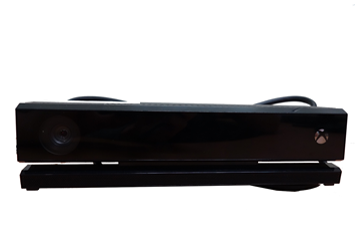}}
    \subfigure[]
    {\includegraphics[width=0.31\linewidth]{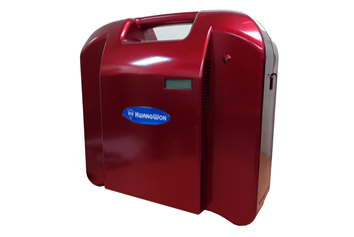}}
    \subfigure[]
    {\includegraphics[width=0.303\linewidth]{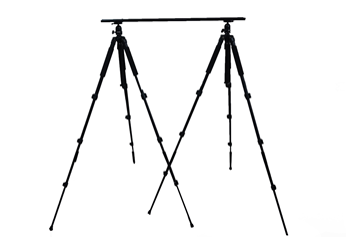}}\\
    \vspace{-12pt}
    \caption{Hardware setup for indoor data: (from left to right) Microsoft Kinect v2~\cite{kinect}, external battery, and tripod and slider.}
    \label{fig:1}
\end{figure}

We captured RGB-D videos of different indoor scenes using the Microsoft Kinect v2 (time-of-flight sensor)~\cite{kinect}.
It has been known that compared to the Kinect v1 (structured light method), the Kinect v2 provides sharper and more detailed depth maps with high-quality color streams, while it does not work well on slanted surfaces.
We believe that our indoor dataset is complementary to the NYU dataset taken with the Kinect v1~\cite{nyu} in numerous depth related applications.
Fig.~\ref{fig:1} shows the hardware setup for indoor data acquisition.
An external battery was utilized to enhance the mobility of the Kinect v2 in the hand-held setup.
Additionally, our RGB-D data was captured with a tripod and slider as shown in Fig.~\ref{fig:1}. 
The average camera velocity was slow enough (10 frames per second) to avoid motion blur in color images.

\subsubsection{Data Acquisition}
The color and depth streams were originally captured with 1920$\times$1080 and 512$\times$424 resolutions, respectively.
We set a threshold of 0.5$\sim$7 meters due to inaccuracies in depth measurements at near and far ranges.
We also provide additional 0.1M IR images aligned to color images.

\begin{figure}
    \centering
    \renewcommand{\thesubfigure}{}
    \subfigure{\includegraphics[width=0.117\textheight]{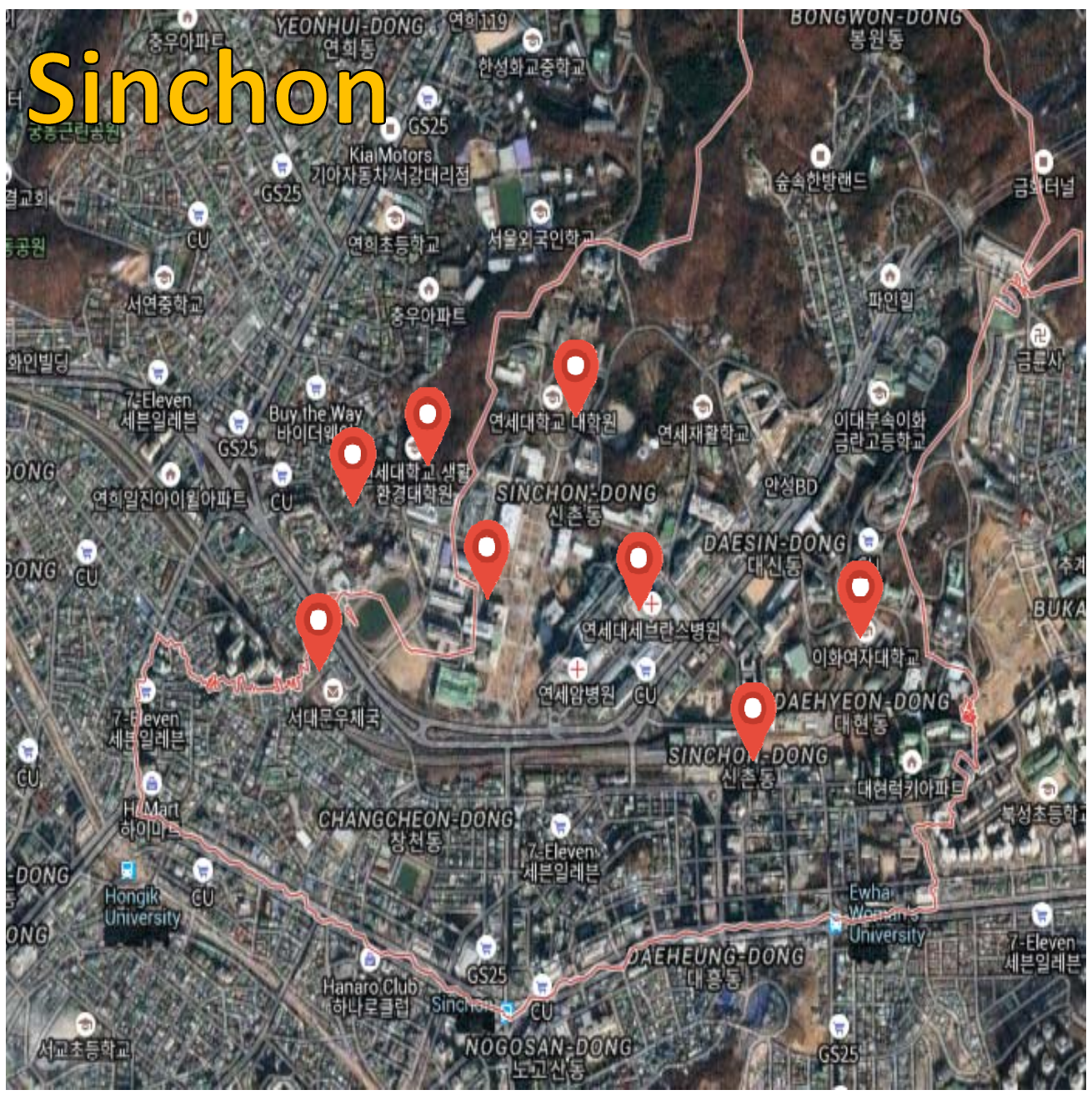}}
    \subfigure{\includegraphics[width=0.117\textheight]{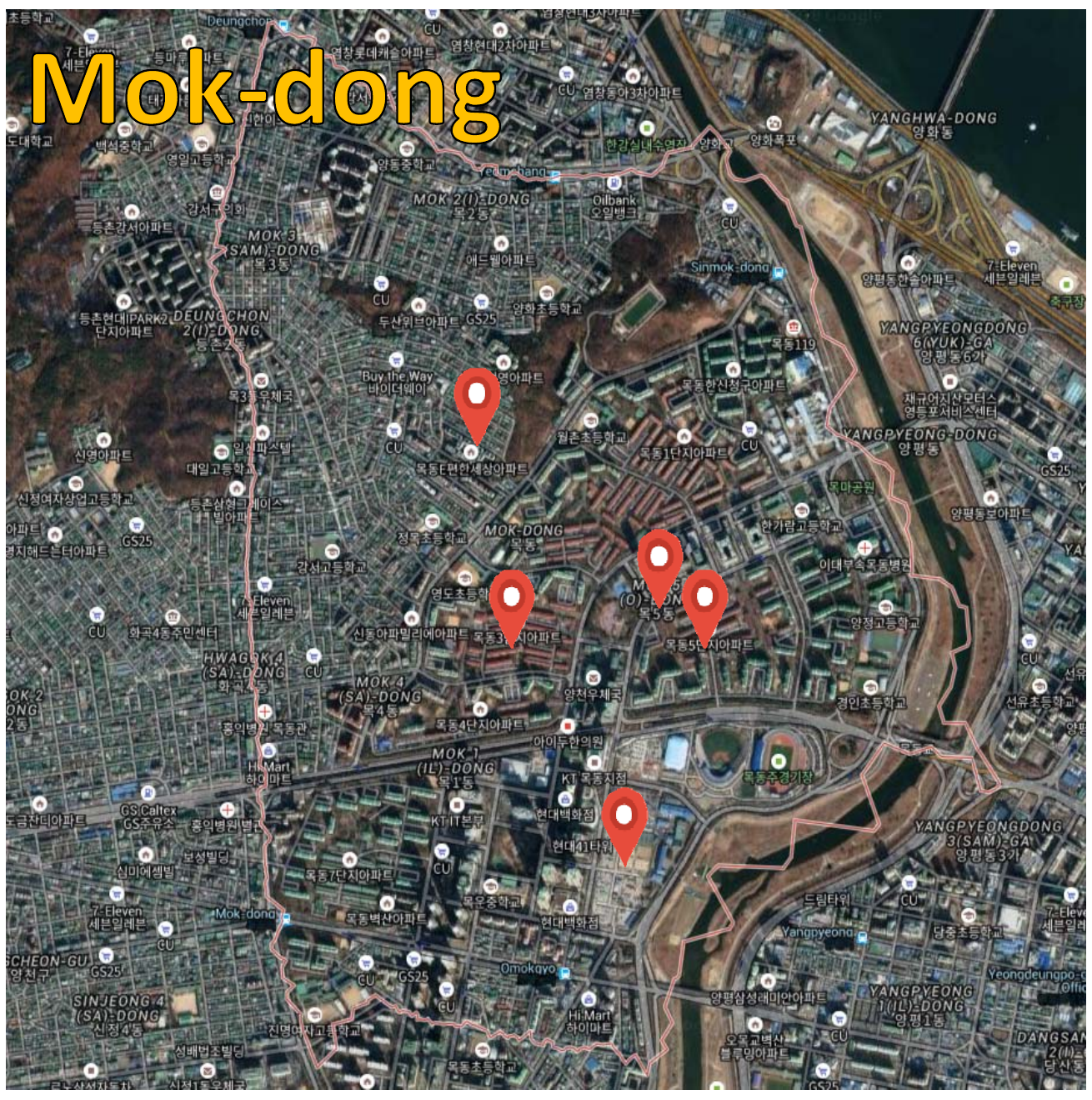}}
    \subfigure{\includegraphics[width=0.117\textheight]{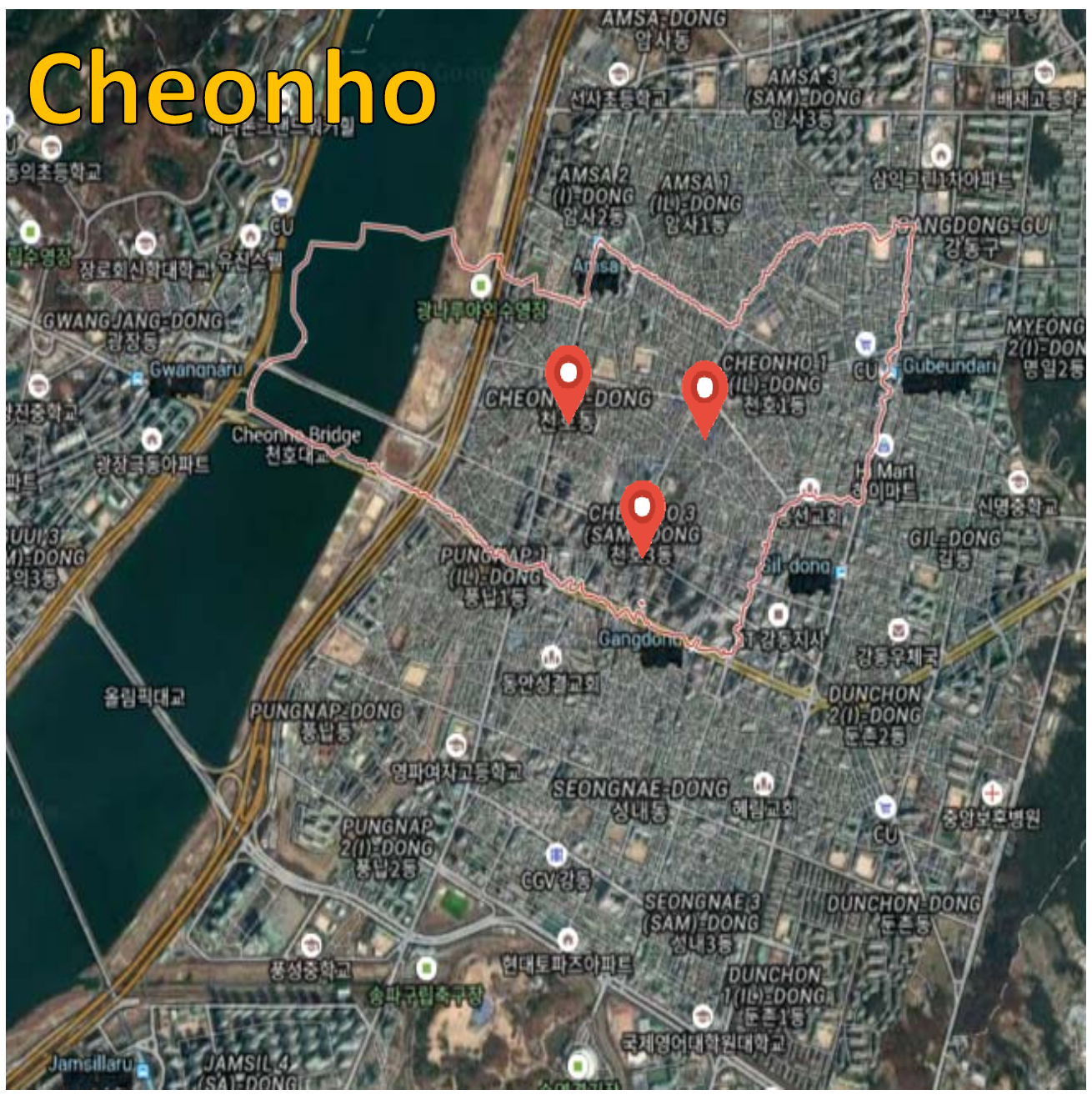}}
    \\ \vspace{-5pt}
    \subfigure{\includegraphics[width=0.117\textheight]{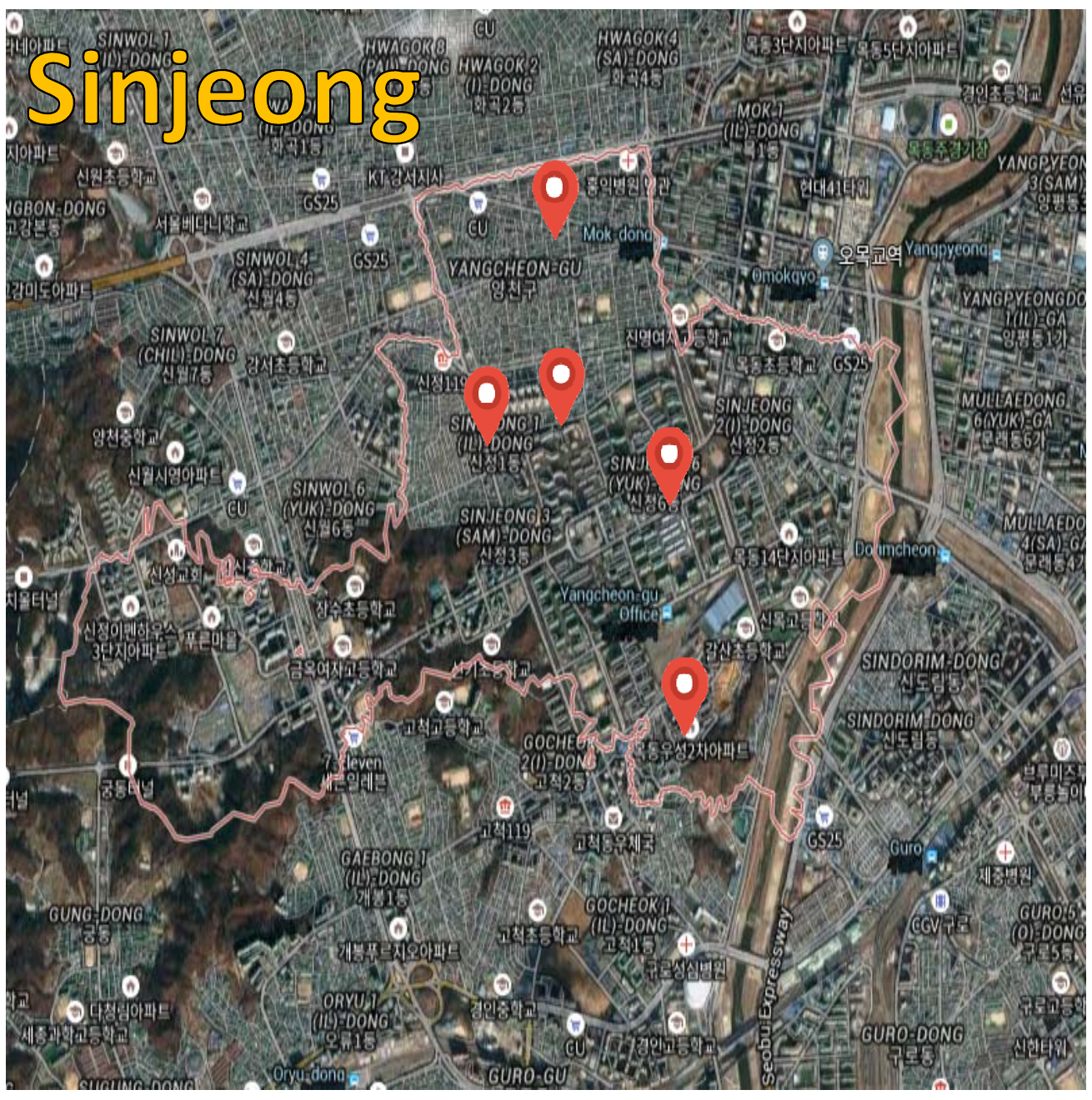}}
    \subfigure{\includegraphics[width=0.117\textheight]{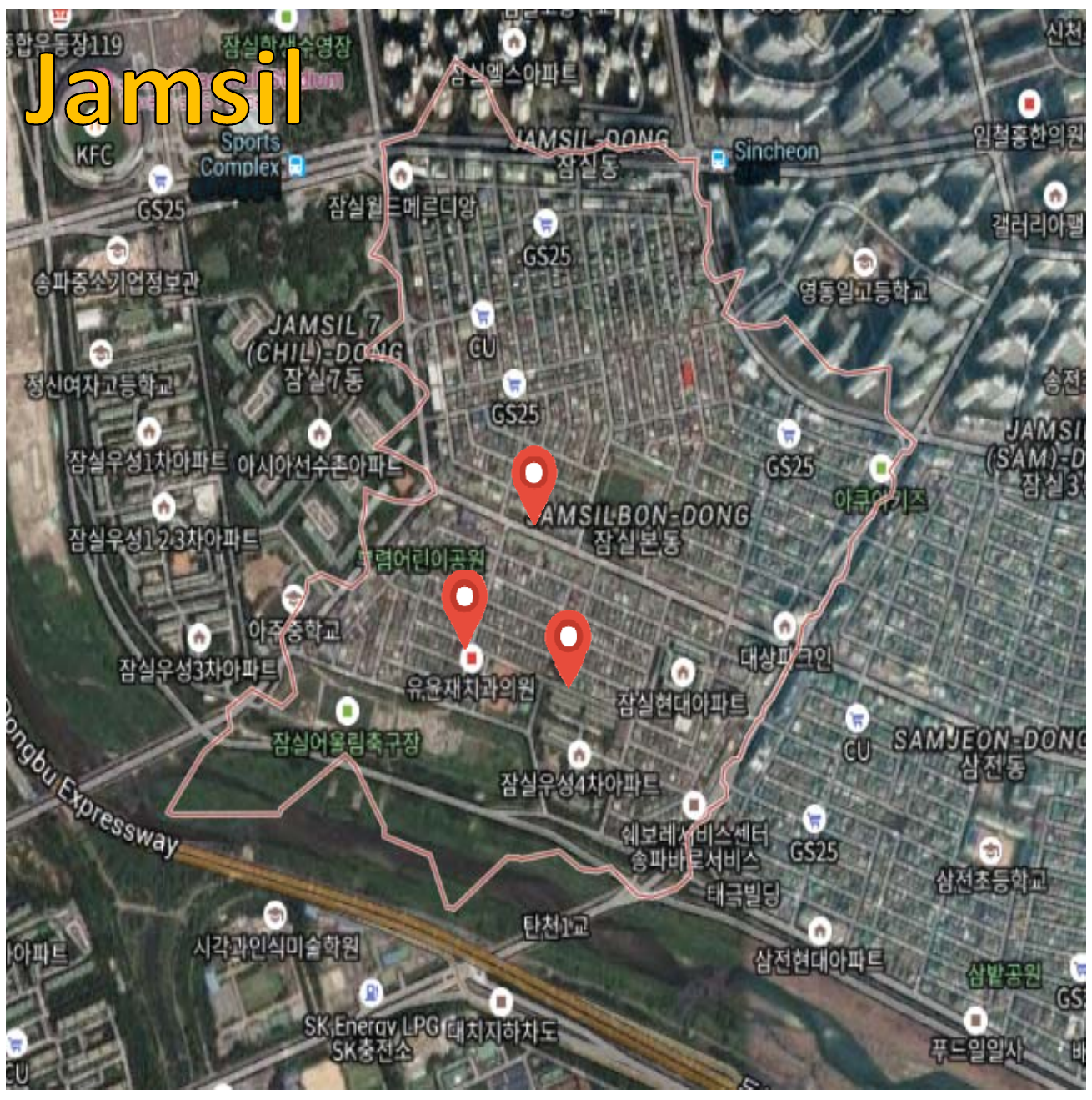}}
    \subfigure{\includegraphics[width=0.117\textheight]{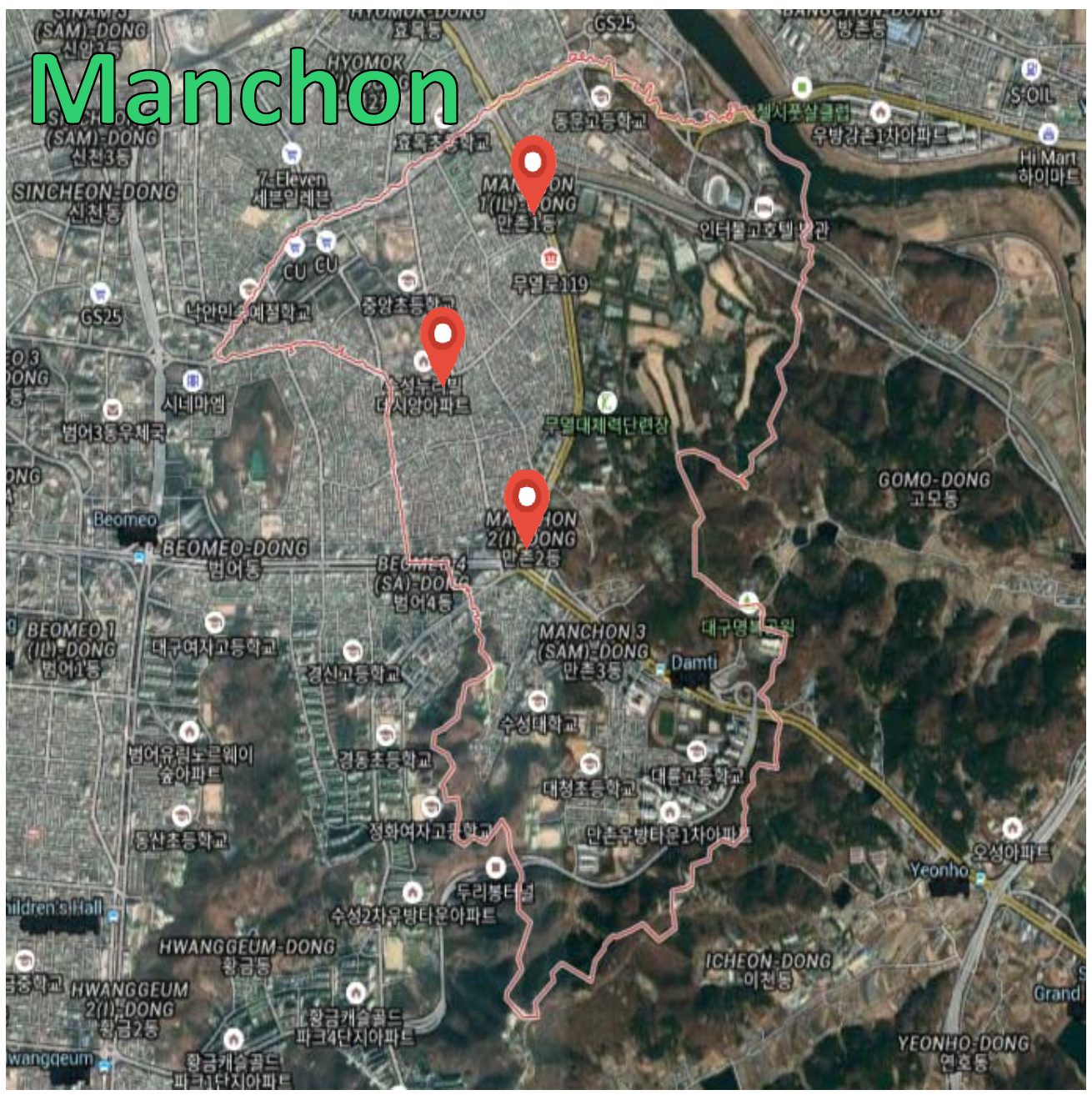}}
    \\ \vspace{-5pt}
    \subfigure{\includegraphics[width=0.117\textheight]{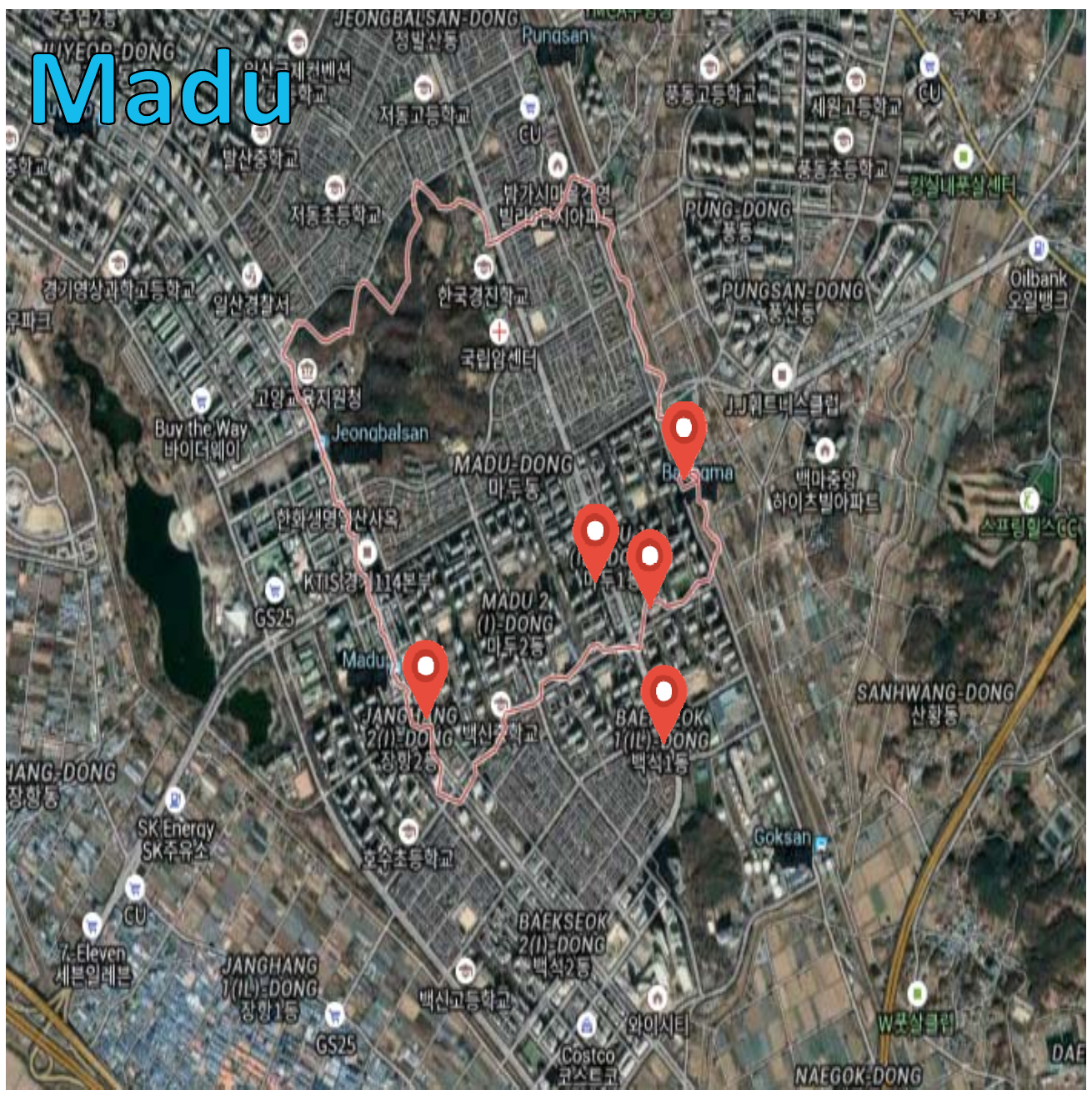}}
    \subfigure{\includegraphics[width=0.117\textheight]{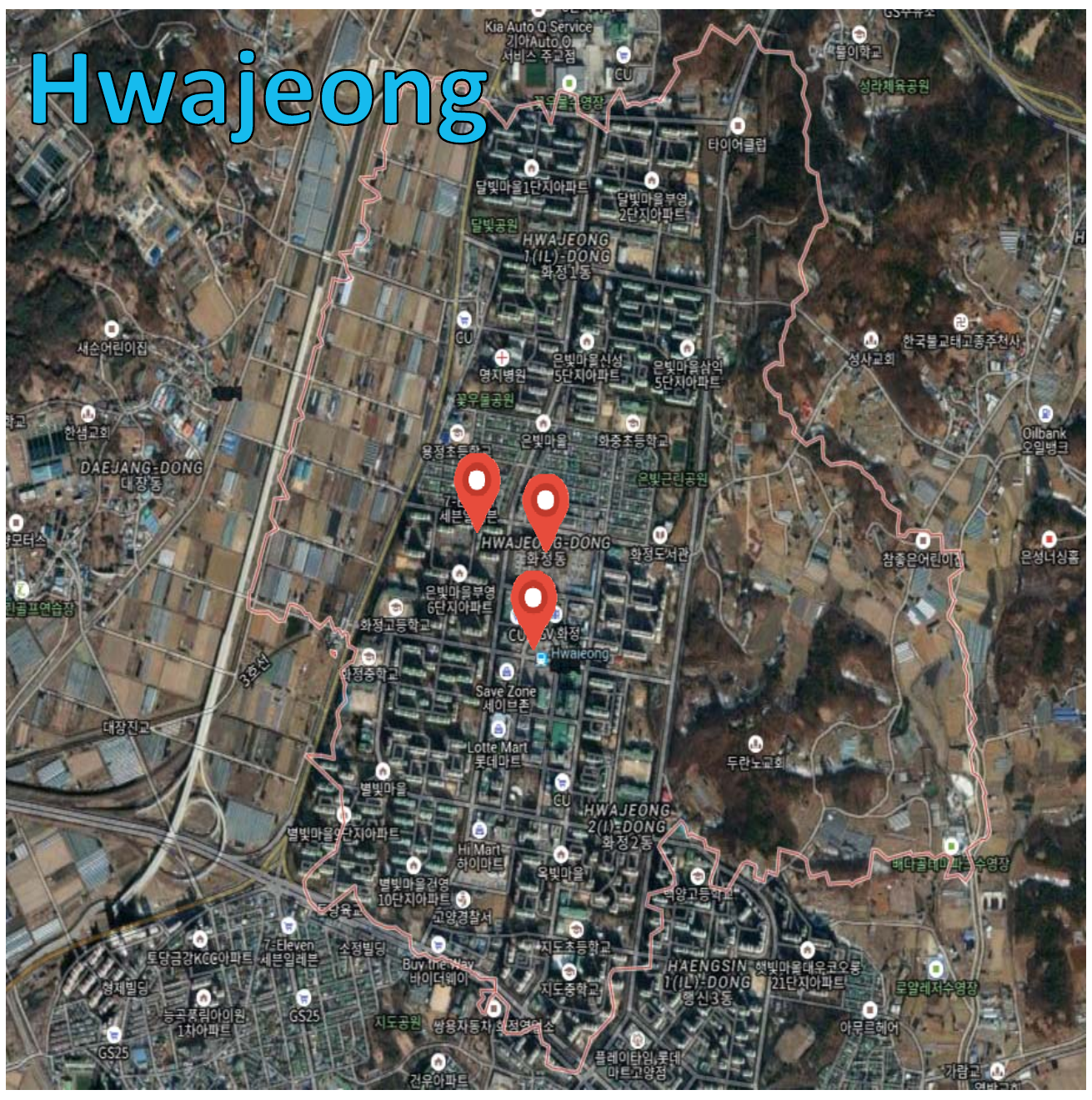}}
    \subfigure{\includegraphics[width=0.117\textheight]{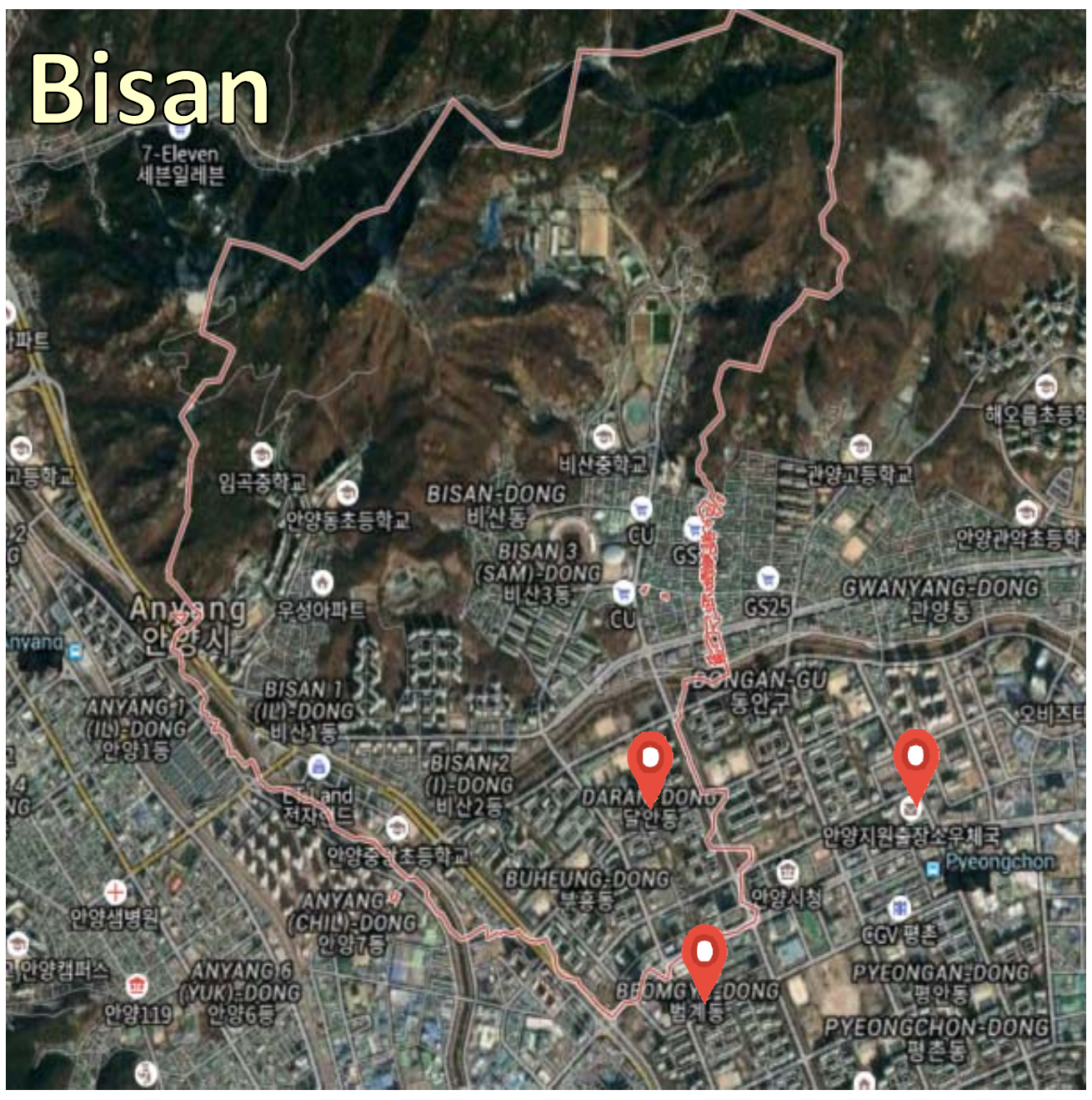}}
    \\
    \caption{Shooting locations of indoor data. Our indoor scenes were captured at various places, e.g., offices, rooms, dormitory, exhibition center, street, and so on. These are located in different 9 districts of 4 cities (South Korea).}
    \label{fig:7}
\end{figure}

Our indoor scenes include a variety of commercial buildings and residential areas including office, room, exhibition center, to name just a few,
which are located in different 9 districts of 4 cities (Seoul, Daegu, Goyang, and Anyang), as shown in Fig.~\ref{fig:7}.
The total numbers of categories and scenes are 18 and 283, as summarized in Table~\ref{tab:4}. 

The categories include warehouse, cafe, classroom, church, computer room, meeting room, library, laboratory, bookstore, corridor, bedroom, living room, kitchen, bathroom, restaurant, billiard, hospital, and store.
Especially, ``bedroom" and ``living room" contain a variety of scenes in indoor interior exhibition halls.

\begin{figure*}
    \centering
    \renewcommand{\thesubfigure}{}
    \subfigure[]
    {\includegraphics[width=0.9\linewidth]{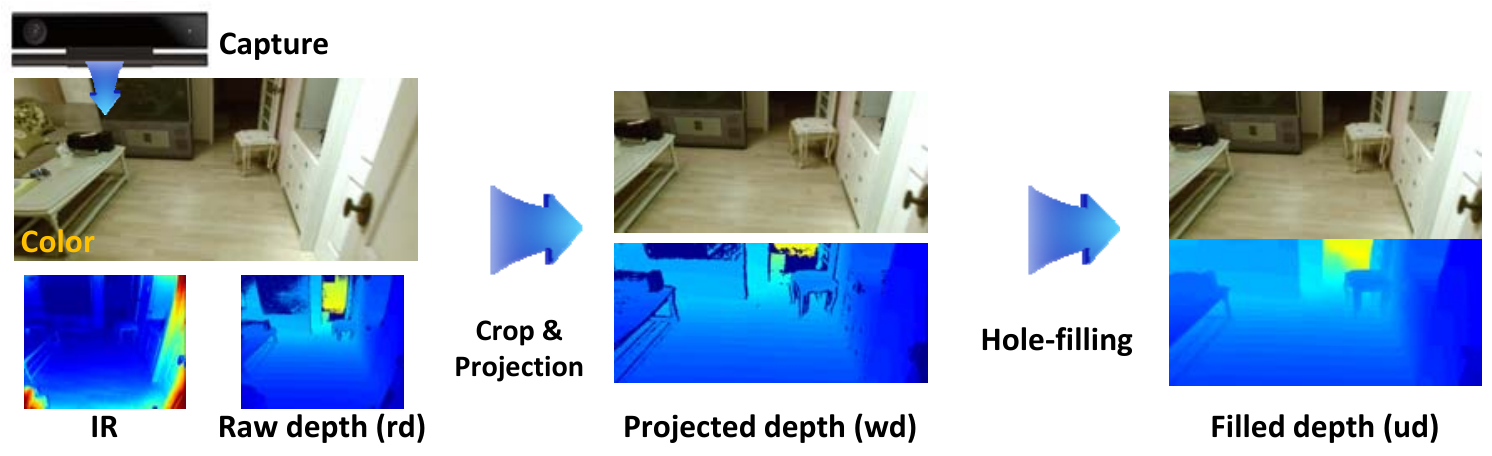}}\\
    \vspace{-3pt}
    \caption{The processing pipeline for our indoor dataset.}
    \label{fig:2}
\end{figure*}

\subsection{Data Processing}
{The overall pipeline for processing indoor data is illustrated in Fig.~\ref{fig:2}.
To co-align color image and depth map, we first projected the raw depth map into the RGB camera coordinate.
The holes in the warped depth map were then filled using color-guided depth upsampling algorithm.}

\subsubsection{Calibration}
We calibrated and aligned the indoor data using 1) Kinect SDK or 2) `iai$\_$kinect v2 libary'~\cite{iai}.
Approximately 476,000 image pairs (taken from fall 2015 to summer 2016) out of 1M RGB-D data were processed using the MS Kinect SDK.
A depth map was projected into the RGB camera coordinate using the Kinect SDK built-in functions.

\begin{figure}[t]
	
     \centering
	 \renewcommand{\thesubfigure}{}
	 \subfigure{\includegraphics[width=0.285\linewidth]{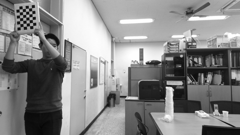}}
	 \subfigure{\includegraphics[width=0.192\linewidth]{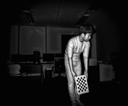}}
	 \subfigure{\includegraphics[width=0.285\linewidth]{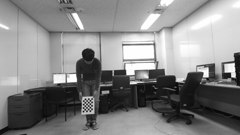}}
	 \subfigure{\includegraphics[width=0.192\linewidth]{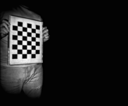}}	 
	 \\ \vspace{-5pt}
	 \subfigure{\includegraphics[width=0.285\linewidth]{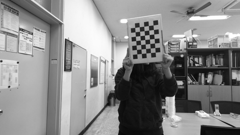}}
	 \subfigure{\includegraphics[width=0.192\linewidth]{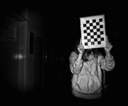}}
	 \subfigure{\includegraphics[width=0.285\linewidth]{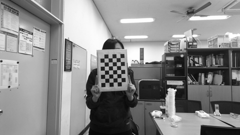}}
	 \subfigure{\includegraphics[width=0.192\linewidth]{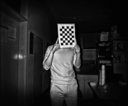}}
	 \\ \vspace{-5pt}
	 \subfigure{\includegraphics[width=0.285\linewidth]{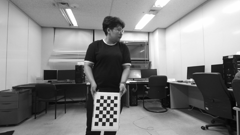}}
	 \subfigure{\includegraphics[width=0.192\linewidth]{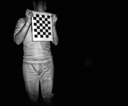}}
	 \subfigure{\includegraphics[width=0.285\linewidth]{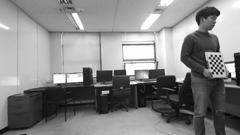}}
	 \subfigure{\includegraphics[width=0.192\linewidth]{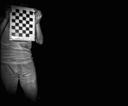}}
	 \\ \vspace{-5pt}
	 \caption{Calibration board images for indoor data. We obtained several images by tilting the calibration pattern vertically and horizontally. At least 150 images were taken so that calibration pattern covers in all areas of the image.}
	 \label{fig:3}	 
\end{figure}

The `iai kinect v2 libary' was used to process the remaining RGB-D image pairs (taken from fall 2016 to summer 2017).
We acquired several pieces of color, IR, and raw depth images containing a checkerboard pattern of 5$\times$7 as shown in Fig. \ref{fig:3}.
Here, the distance between corners was set to 0.03m.
Note that for an accurate calibration, the pattern images should be taken in a way of covering most of the image at various ranges.
Using the calibration toolbox provided in the `iai kinect v2 libary'~\cite{iai}, we estimated the shift parameter between IR and depth images and the projection matrix between IR and color images.
The warping process consists of two parts. The depth map was first warped into IR image coordinate using a shift parameter.
The depth map on IR camera coordinate was finally projected into the RGB image coordinate.

Note that warped depth maps has two different sizes depending on calibration toolbox used.
The MS Kinects SDK~\cite{kinect} warped the depth map into the RGB image coordinate of 1/4 spatial resolution ($480\times 270$),
while the `iai kinect v2 libary'~\cite{iai} performed the warping into the cropped RGB image coordinate.
After projecting depth values to the RGB camera coordinate, we discarded the region exceeding the field of view of IR camera.
Thus, the cropped color image is of 1408$\times$792.

\subsubsection{Depth upsampling}
We applied the color-guided depth upsampling algorithm~\cite{Kim2016} to fill the holes in depth maps.

\subsection{Data Format}
We stored depth maps in mm using a 16bit portable network graphics (png) format.

The indoor dataset includes the following files. Here, ``{\bf{in}}" denotes the indoor data.\\

\begin{itemize}
    \item Color image
    \\ ``{\bf{in$\_$k$\_$14-digits$\_$c.png}}"
    \\ 24bit cropped RGB image (1408$\times$792)
    \\
    \item Depth map
    \\ ``{\bf{in$\_$k$\_$14-digits$\_$rd.png}}"
    \\ 16bit original depth map in mm (512$\times$424)
    \vspace{0.1cm}
    \\ ``{\bf{in$\_$k$\_$14-digits$\_$wd.png}}"
    \\ 16bit projected depth map in mm (480$\times$270),(1408$\times$792)
    \vspace{0.1cm}
     \\ ``{\bf{in$\_$k$\_$14-digits$\_$ud.png}}"
     \\ 16bit filled depth map in mm (1408$\times$792)
    \\
    \item Calibration parameters
    \\ ``{\bf{calib$\_$color$\_$2-digits.mat}}"
    \\ intrinsic parameter for color camera
    \vspace{0.1cm}
    \\ ``{\bf{calib$\_$ir$\_$2-digits.mat}}"
    \\ intrinsic parameter for IR camera
    \vspace{0.1cm}
    \\ ``{\bf{calib$\_$depth$\_$2-digits.mat}}"
    \\ shift parameter between IR and depth
    \vspace{0.1cm}
    \\ ``{\bf{calib$\_$pose$\_$2-digits.mat}}"
    \\ extrinsic parameter between color and IR cameras
\end{itemize}

A total of four Kinect v2 cameras were used, and we assigned an unique ID (14 digits) to each RGB-D image pair.
The ID consists of 2 digits for camera number ($01\sim04$), 6 digits for date (year, month, day), and 6 digits for order of acquisition , respectively (for example, `in$\_$k$\_$01$\_$160220$\_$000001').

\begin{table}[t]
    \centering
    \caption{Indoor dataset categorization.
     We uploaded the data acquired using SDK and we will post our own calibration data later.}
    \vspace{-5pt}
    \resizebox{0.4\textwidth}{!}{
    \begin{tabular}{ccccc}
        \toprule
        {\bf }                     & {\bf Category}       & {\bf  $\#$ of folders} & {\bf $\#$ of files} \\ \hline
        \midrule
        \raisebox{-\totalheight}{\includegraphics[width=0.2\textwidth]{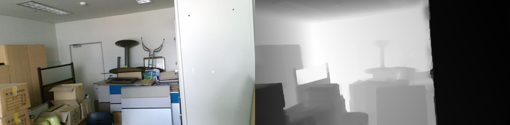}} & \multirow{4}{*}{Warehouse} & \multirow{4}{*}{26}  & \multirow{4}{*}{33923} \\
        \raisebox{-\totalheight}{\includegraphics[width=0.2\textwidth]{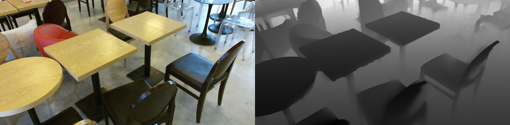}} & \multirow{4}{*}{Cafe}      & \multirow{4}{*}{22}  & \multirow{4}{*}{33515}       \\
        \raisebox{-\totalheight}{\includegraphics[width=0.2\textwidth]{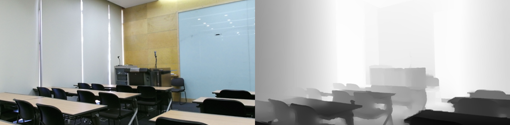}}& \multirow{4}{*}{Classroom} & \multirow{4}{*}{65} & \multirow{4}{*}{95867}               \\
        \raisebox{-\totalheight}{\includegraphics[width=0.2\textwidth]{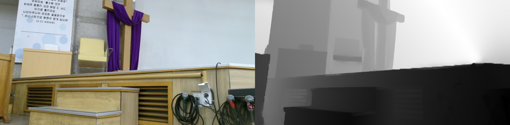}}      & \multirow{4}{*}{Church} & \multirow{4}{*}{10}  & \multirow{4}{*}{18161}               \\
        \raisebox{-\totalheight}{\includegraphics[width=0.2\textwidth]{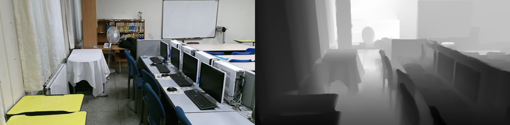}}    & \multirow{4}{*}{Computer Room} & \multirow{4}{*}{7}  & \multirow{4}{*}{12583}               \\
        \raisebox{-\totalheight}{\includegraphics[width=0.2\textwidth]{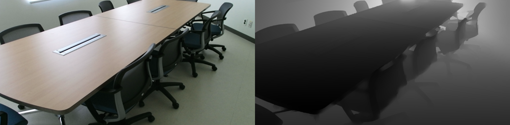}}  & \multirow{4}{*}{Meeting Room} & \multirow{4}{*}{23}  & \multirow{4}{*}{42241}              \\
        \raisebox{-\totalheight}{\includegraphics[width=0.2\textwidth]{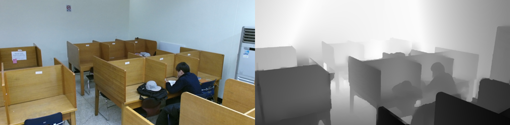}}  & \multirow{4}{*}{Library} & \multirow{4}{*}{15}  & \multirow{4}{*}{22001}               \\
        \raisebox{-\totalheight}{\includegraphics[width=0.2\textwidth]{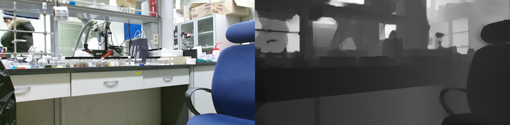}}      & \multirow{4}{*}{Laboratory} & \multirow{4}{*}{22}  & \multirow{4}{*}{29748}               \\
        \raisebox{-\totalheight}{\includegraphics[width=0.2\textwidth]{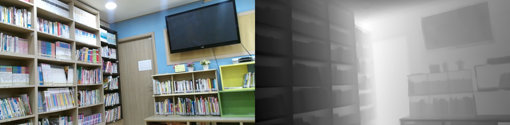}}   & \multirow{4}{*}{Bookstore} & \multirow{4}{*}{5}  & \multirow{4}{*}{9539}               \\
        \raisebox{-\totalheight}{\includegraphics[width=0.2\textwidth]{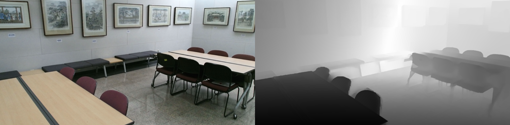}}   & \multirow{4}{*}{Corridor} & \multirow{4}{*}{32}  & \multirow{4}{*}{26236}               \\
        \raisebox{-\totalheight}{\includegraphics[width=0.2\textwidth]{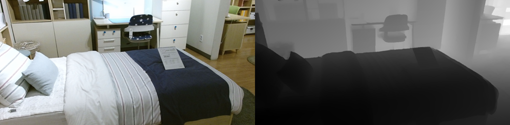}}     & \multirow{4}{*}{Bedroom} & \multirow{4}{*}{33}  & \multirow{4}{*}{55215}               \\
        \raisebox{-\totalheight}{\includegraphics[width=0.2\textwidth]{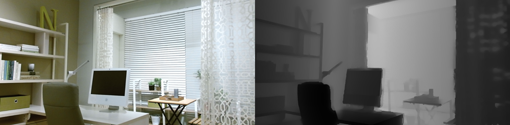}}  & \multirow{4}{*}{Livingroom} & \multirow{4}{*}{28}  & \multirow{4}{*}{49502}             \\
        \raisebox{-\totalheight}{\includegraphics[width=0.2\textwidth]{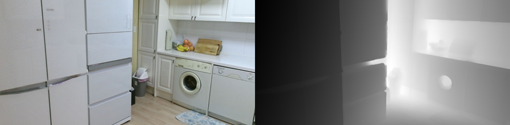}}     & \multirow{4}{*}{Kitchen} & \multirow{4}{*}{6}  & \multirow{4}{*}{9274}              \\
        \raisebox{-\totalheight}{\includegraphics[width=0.2\textwidth]{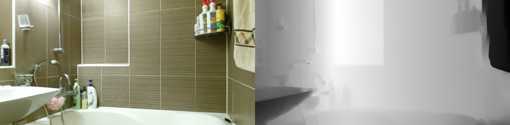}}       & \multirow{4}{*}{Bathroom} & \multirow{4}{*}{1}  & \multirow{4}{*}{1140}              \\
        \raisebox{-\totalheight}{\includegraphics[width=0.2\textwidth]{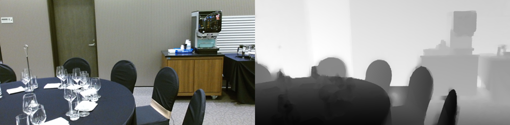}}  & \multirow{4}{*}{Restaurant} & \multirow{4}{*}{22}  & \multirow{4}{*}{33034}        \\
        \raisebox{-\totalheight}{\includegraphics[width=0.2\textwidth]{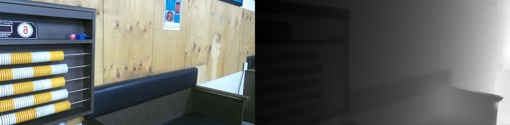}}   & \multirow{4}{*}{BilliardHall} & \multirow{4}{*}{1}  & \multirow{4}{*}{3523}       \\
        \raisebox{-\totalheight}{\includegraphics[width=0.2\textwidth]{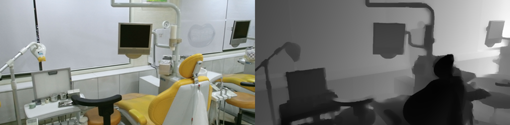}}  & \multirow{4}{*}{Hospital} & \multirow{4}{*}{0}  & \multirow{4}{*}{0}           \\
        \raisebox{-\totalheight}{\includegraphics[width=0.2\textwidth]{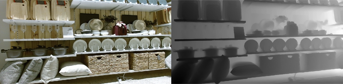}}       & \multirow{4}{*}{Store} & \multirow{4}{*}{5}  & \multirow{4}{*}{5606}              \\
        \bottomrule
    \end{tabular}}\label{tab:4}
\end{table}

\subsection{Indoor data toolbox}
\begin{itemize}
    \item ``{\bf{fill$\_$depth.m}}" - Fills in the holes in depth map.    
\end{itemize}
Refer to~\cite{kinect} and~\cite{iai} about projecting the depth map and cropping the color region exceeding the field of view of IR camera.

\section{Outdoor RGB-D dataset}
Unlike indoor scenes, obtaining an accurate and dense depth map of outdoor scenes poses additional challenges due to strong illumination and longer depth range, making it infeasible to use the time-of-flight sensor (Kinect v2).
LiDAR can be an alternative for depth acquisition as in the KITTI dataset~\cite{kitti}, but it produces a sparse depth map, e.g., for less than 6$\%$ of pixels in 1224$\times$386 resolution.
Aside from this, depth values are provided only at the bottom part of an color image due to a limited vertical field of view of LiDAR (typically ${40^ \circ }$).
To increase the density of the captured depth maps, we may accumulate multiple sparse depth maps (from LiDAR) using the iterative closest point (ICP) algorithm~\cite{Segal09}.
However, significant manual effort is required to remove outliers and artifacts in the depth map, due to the different centers of projection of the LiDAR and color camera or reflecting/transparent surfaces.
Thus, this approach is not suitable for acquiring large scale RGB-D data in outdoors.

Alternatively, dense ground truth depth maps would be generated through synthetic rendering as in~\cite{Mayer16}, but they incur the domain adaptation issue inherently when directly used for natural outdoor scenes.

Considering these factors, we adopted the stereo based approach for generating large scale RGB-D data in outdoor scenes.
The depth map is computed using stereo matching algorithms~\cite{Zbontar2015}.
To minimize the side effects of incorrectly estimated depth values, we additionally provide a confidence map~\cite{Park2015} and~\cite{Kim2018} associated with the estimated depth map.

\subsection{Data Acquisition}
\subsubsection{Hardware setup}
\begin{figure}
    \centering
    \renewcommand{\thesubfigure}{}
    \subfigure[]
    {\includegraphics[width=0.303\linewidth]{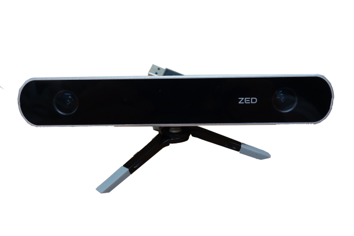}}
    \subfigure[]
    {\includegraphics[width=0.288\linewidth]{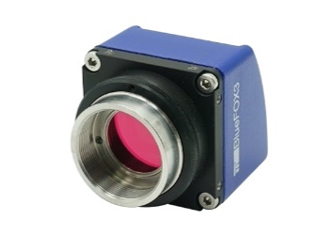}}
    \subfigure[]
    {\includegraphics[width=0.37\linewidth]{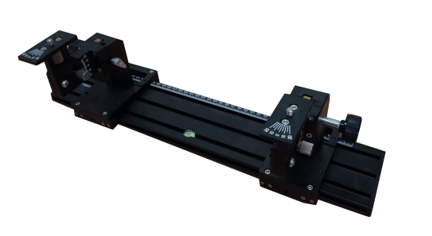}}\\
    \vspace{-15pt}
    \caption{Hardware setup for outdoor data: (from left to right) ZED stereo camera~\cite{zed}, mvBlueFox3 sensor~\cite{fox}, and slider.}
    \label{fig:4}
\end{figure}

We used the ZED camera and built-in stereo camera for acquiring stereo images (Fig. \ref{fig:4}).
Although the commercial ZED camera provide high quality stereo images, the sensing range is rather limited (up to 20m according to the specification sheet) due to a relatively small baseline (12cm).
We designed the built-in stereo system with mvBlueFox3 sensors~\cite{fox} and set the baseline to 40cm, increasing the sensing range up to 80m.
The mvBlueFox3 SDK~\cite{fox} was used for synchronization.
The devices used for outdoor data acquisition are shown in Fig. \ref{fig:4}.

\begin{figure}
    \centering
    \renewcommand{\thesubfigure}{}
    \subfigure{\includegraphics[width=0.16\textheight]{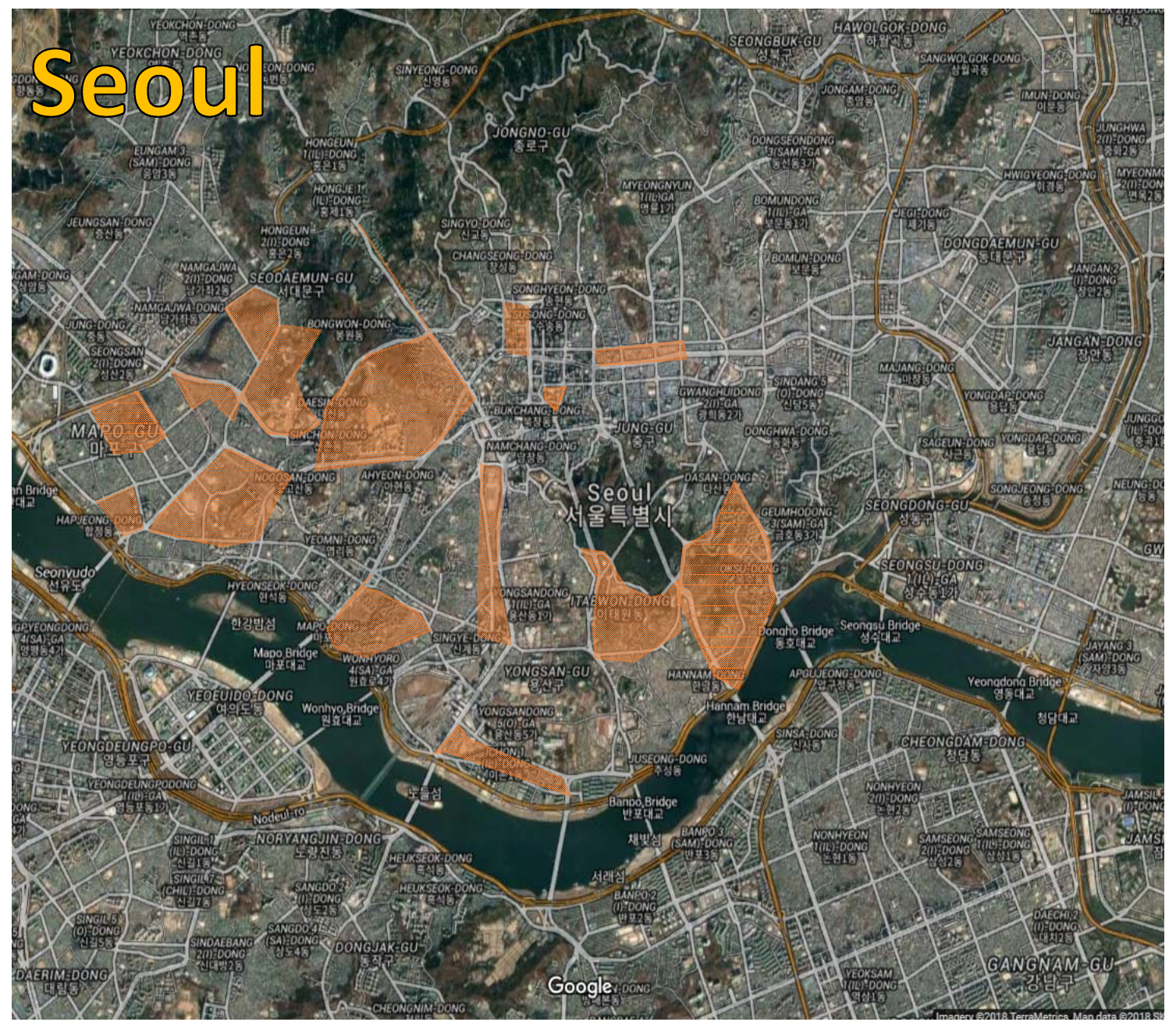}}
    \subfigure{\includegraphics[width=0.16\textheight]{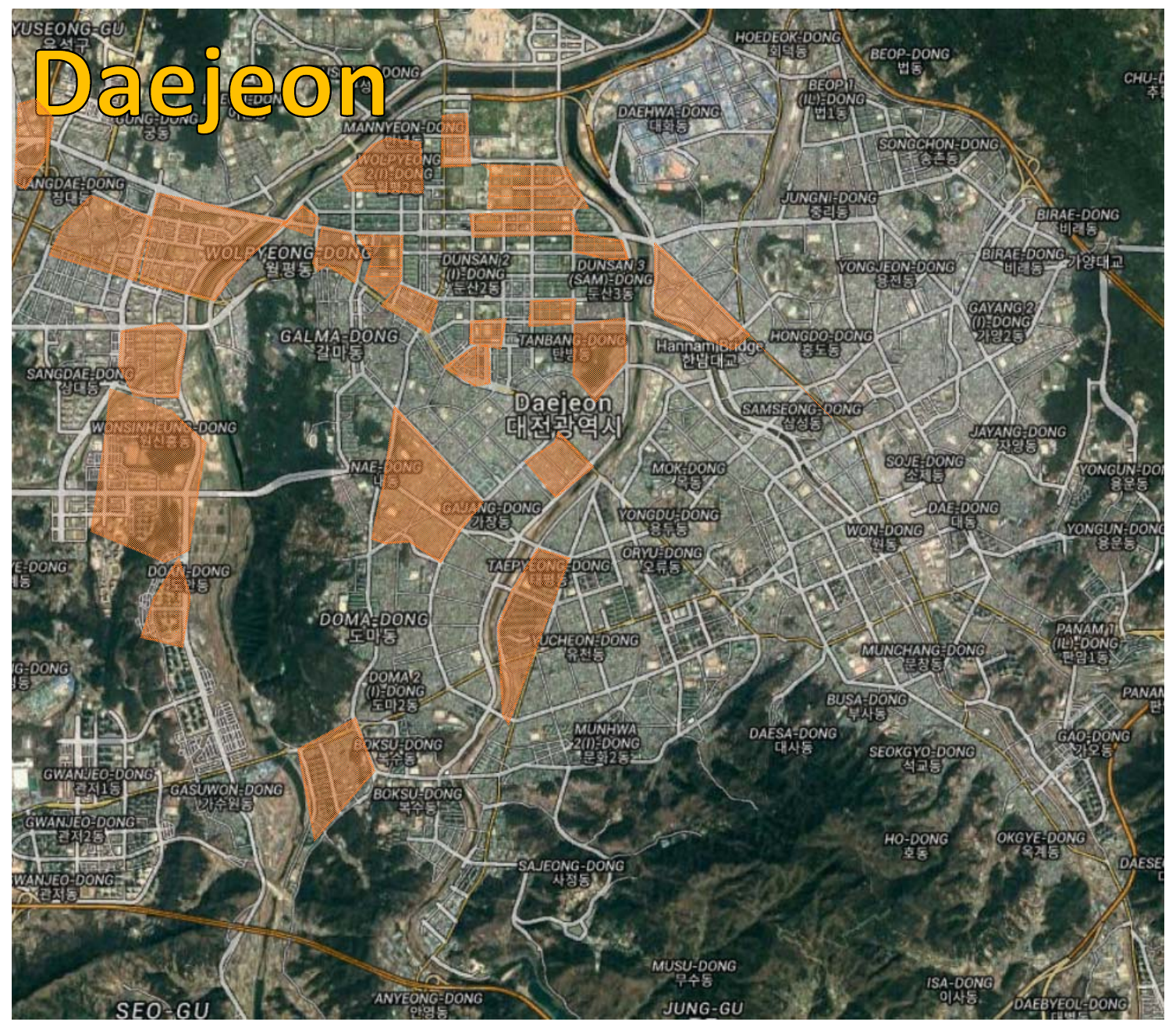}} \\\vspace{-5pt}
    \subfigure{\includegraphics[width=0.16\textheight]{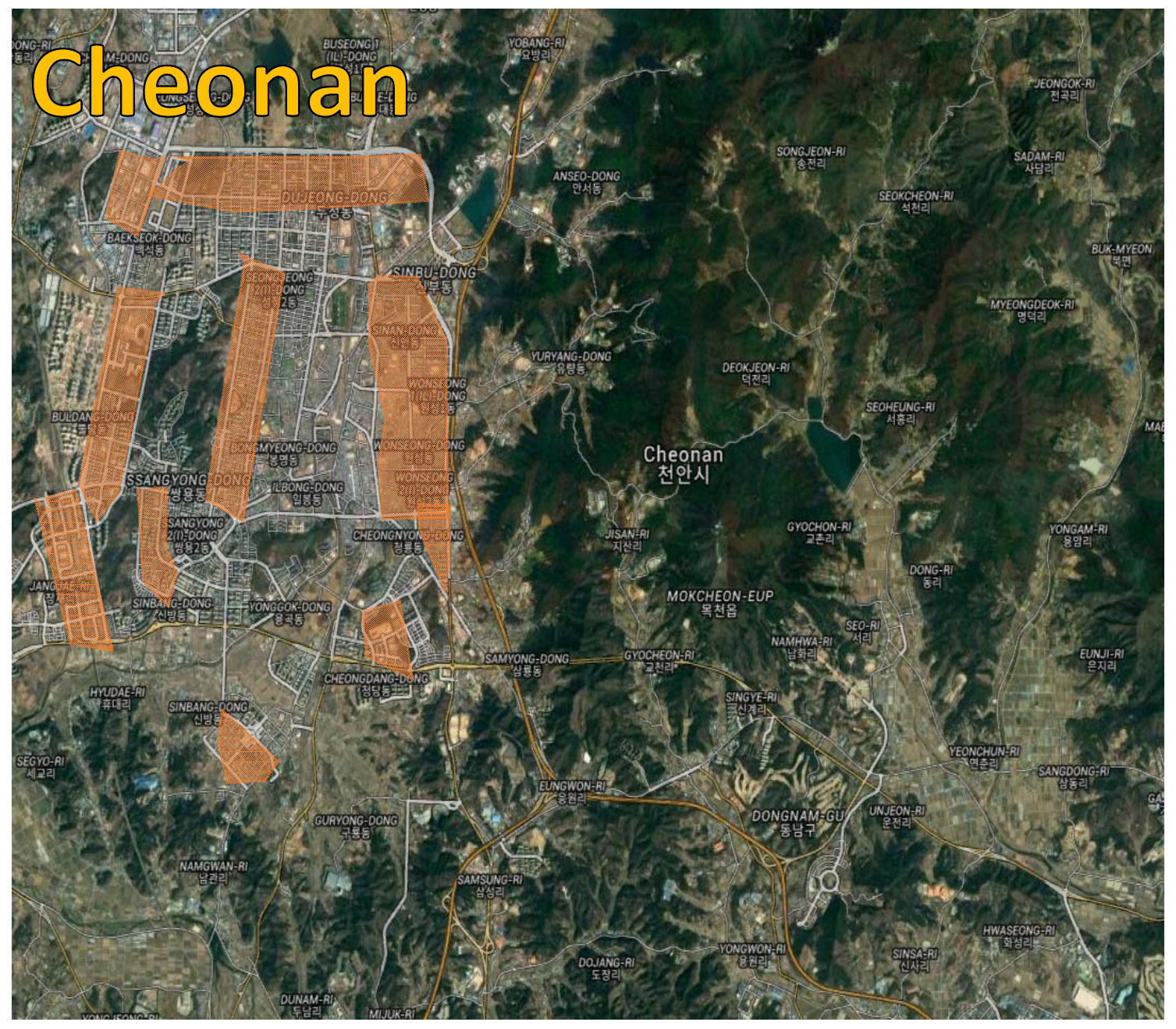}}
    \subfigure{\includegraphics[width=0.16\textheight]{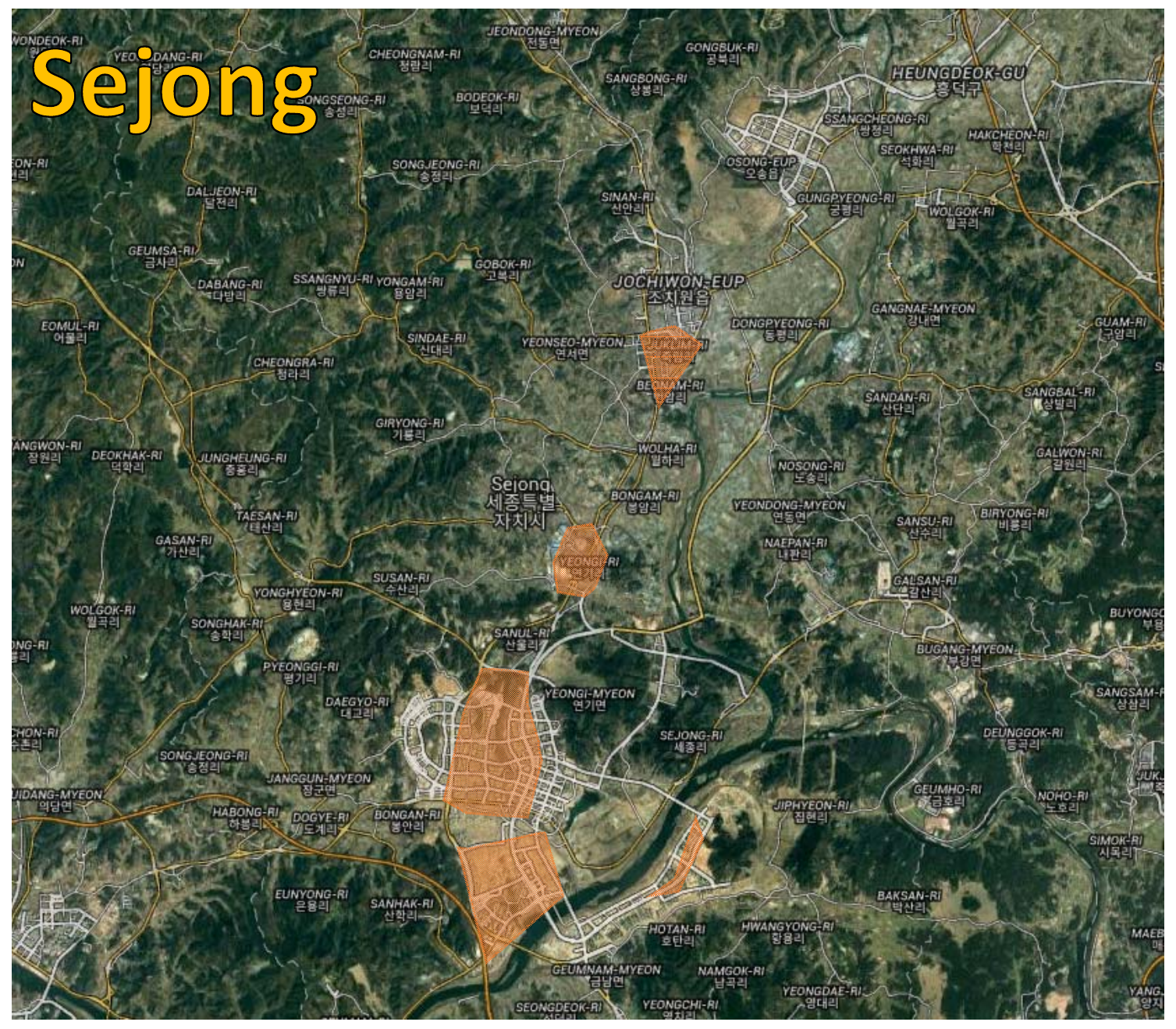}}
    \\
    \caption{Shooting locations of outdoor data. Our dataset acquired various outdoor scenes using the hand-held device, including park, building, brook, road, apartment, and so on. These are located in different 4 cities in South Korea.}
    \label{fig:8}
\end{figure}

\begin{figure*}
    \centering
    \renewcommand{\thesubfigure}{}
    \subfigure[]
    {\includegraphics[width=0.97\linewidth]{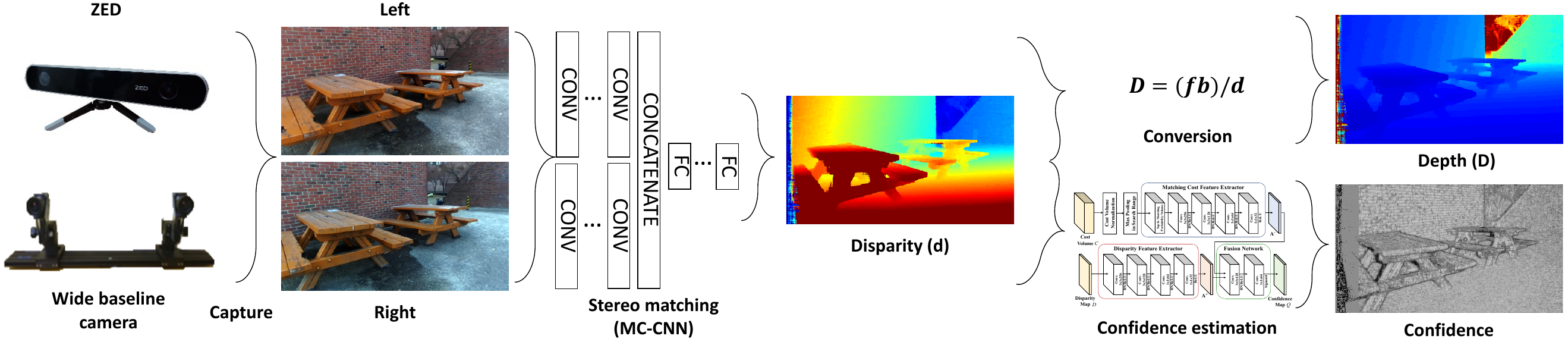}}\\
    \vspace{-12pt}
    \caption{The processing pipeline for our outdoor dataset. We captured stereo images, and then generated depth maps using the MC-CNN stereo matching~\cite{Zbontar2015}, as well as per-pixel confidence maps.}
    \label{fig:5}
\end{figure*}
    \vspace{-3pt}
    
\subsubsection{Data acquisition}
The stereo image was captured with 1920$\times$1080 or 1280$\times$720 resolutions.
0.92M and 0.08M stereo images were obtained using the ZED and the built-in stereo camera, respectively.
The average camera velocity was 10 frames per second to avoid motion blur in stereo images.

The stereo images of the KITTI dataset~\cite{kitti} were obtained through stereo camera mounted on a moving vehicle, and thus the outdoor data is mainly comprised of road and traffic scenes.
In contrast, our dataset acquired various outdoor scenes using the hand-held device, including park, building, brook, road, apartment, and so on, which are located in different 4 cities (Seoul, Daejeon, Cheonan, and Sejong), as shown in Fig. \ref{fig:8}.
The total numbers of category and scene are 9 and 132, as summarized in Table \ref{tab:5}.
All the scenes were taken steadily with a tripod and slider while walking.
We believe that our outdoor dataset is complementary to the existing outdoor datasets thanks to the diverse scene coverage.

\begin{table}
    \centering
    \caption{Outdoor dataset categorization. We uploaded the data acquired with~\cite{Park2015} using ZED and we will post it later.}
    \resizebox{0.45\textwidth}{!}{
    \begin{tabular}{cccc}
        \toprule
        {\bf } & {\bf Category} & {\bf$\#$ of folders} & {\bf$\#$ of files} \\ \hline
        \midrule
        \raisebox{-\totalheight}{\includegraphics[width=0.2\textwidth]{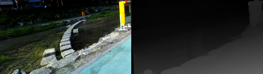}}
        & \multirow{4}{*}{brook} & \multirow{4}{*}{1}  & \multirow{4}{*}{4672}                \\
        \raisebox{-\totalheight}{\includegraphics[width=0.2\textwidth]{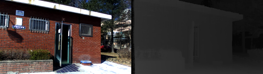}}
        & \multirow{4}{*}{building} & \multirow{4}{*}{22}  & \multirow{4}{*}{58704}              \\
        \raisebox{-\totalheight}{\includegraphics[width=0.2\textwidth]{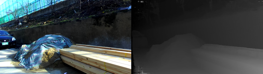}}
        & \multirow{4}{*}{construction} & \multirow{4}{*}{1} & \multirow{4}{*}{1871}           \\
        \raisebox{-\totalheight}{\includegraphics[width=0.2\textwidth]{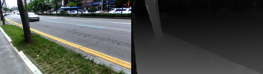}}
        & \multirow{4}{*}{driveway} & \multirow{4}{*}{7}  & \multirow{4}{*}{11114}               \\
        \raisebox{-\totalheight}{\includegraphics[width=0.2\textwidth]{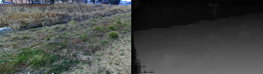}}
        & \multirow{4}{*}{field} & \multirow{4}{*}{3}  & \multirow{4}{*}{3039}               \\
        \raisebox{-\totalheight}{\includegraphics[width=0.2\textwidth]{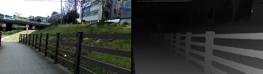}}
        & \multirow{4}{*}{overpass} & \multirow{4}{*}{1}  & \multirow{4}{*}{2794}              \\
        \raisebox{-\totalheight}{\includegraphics[width=0.2\textwidth]{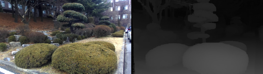}}
        & \multirow{4}{*}{park} & \multirow{4}{*}{10}  & \multirow{4}{*}{23384}               \\
        \raisebox{-\totalheight}{\includegraphics[width=0.2\textwidth]{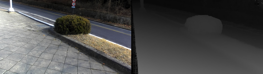}}
        & \multirow{4}{*}{street} & \multirow{4}{*}{75}  & \multirow{4}{*}{198097}               \\
        \raisebox{-\totalheight}{\includegraphics[width=0.2\textwidth]{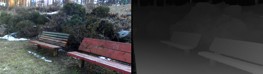}}
        & \multirow{4}{*}{trail} & \multirow{4}{*}{9}  & \multirow{4}{*}{18762}               \\
        \bottomrule
    \end{tabular}}\label{tab:5}
\end{table}

\subsection{Data Processing}
This task involves rectifying the stereo image and estimating a disparity map and its associated confidence map.
Figure \ref{fig:5} illustrates the overall pipeline for processing outdoor data.

\subsubsection{Calibration and Rectification}
\begin{figure}[t]
    \centering
    \renewcommand{\thesubfigure}{}
    \subfigure[]
    {\includegraphics[width=1\linewidth]{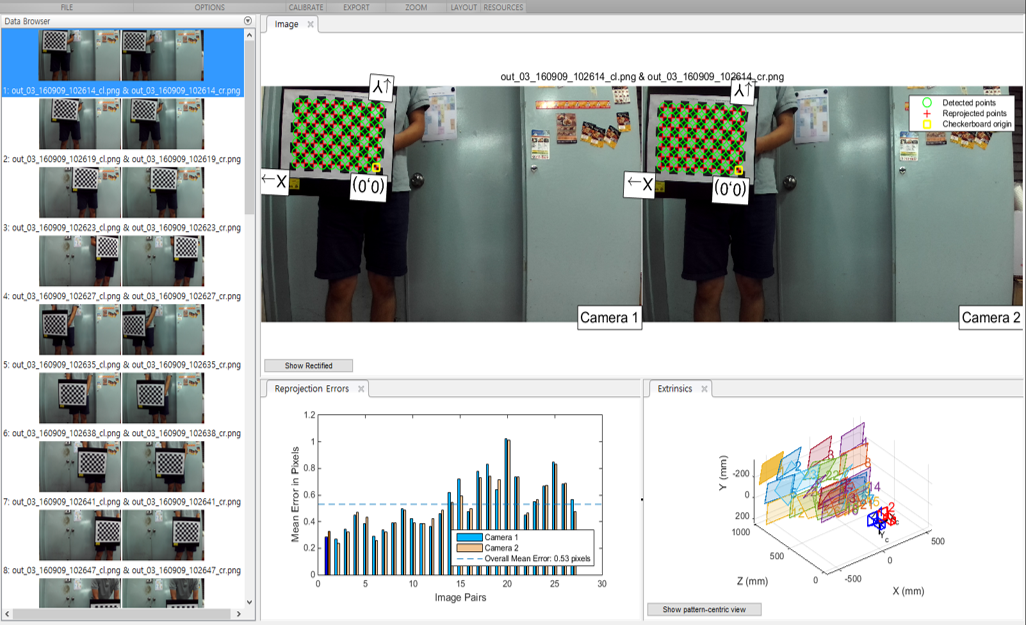}}
    \vspace{-12pt}
    \caption{ Stereo camera calibration. We performed camera distortion correction and stereo image rectification.}
    \label{fig:6}
\end{figure}

We performed our own calibration using the Caltech calibration toolbox~\cite{caltech}.
The 7$\times$8$\times$0.26 and 7$\times$10$\times$0.50 checkerboard patterns were used for the ZED and built-in stereo cameras (Fig. \ref{fig:6}), respectively.
After correcting lens distortion, we rectified stereo images onto a common image plane such that the corresponding points exist on the same horizontal coordinate.

\subsubsection{Stereo Matching}
Given the rectified stereo image pairs, we generate the disparity map of the left image using the MC-CNN~\cite{Zbontar2015}.
Specifically, the cost volume filtering of~\cite{Zhang2009} and semi-global matching method~\cite{Hirschmuller2008} were applied to the raw matching cost volume computed from ~\cite{Zbontar2015}.
We then recovered a depth map using calibration parameters (baseline and forcal length).
A maximum disparity range was set to 228.

\subsubsection{Confidence Estimation}
We used two methods~\cite{Park2015,Kim2018} to obtain the confidence map.
For approximately 400,000 pieces (taken from fall 2015 to summer 2016), the confidence maps were obtained by~\cite{Park2015},
while the remaining image pairs (taken from fall 2016 to summer 2017) were processed by~\cite{Kim2018}.
In~\cite{Kim2018}, we devised a new deep architecture that estimates a stereo confidence.
The confidence map is normalized in [0, 255], and the higher is the better.
We found that with the threshold of 0.9 $\times$ 255, the majority of inaccurate depth values were removed. Refer to~\cite{Park2015} and~\cite{Kim2018} for details.

\subsection{Data Format}
We stored outdoor depth maps in mm using a 16bit png format.
The outdoor dataset includes the following files.
Here, ``{\bf{out}}" denotes the outdoor data.
``{\bf{zed}}", and ``{\bf{wb}}" denote ZED stereo camera, and built-in stereo camera, respectively.\\

\begin{itemize}
    \item Color image (1920$\times$1080 or 1280$\times$720)
    \\ ``{\bf{out$\_$zed(wb)$\_$14-digits$\_$left.png}}"
    \\ 24bit rectified left RGB image
    \vspace{0.1cm}
    \\ ``{\bf{out$\_$zed(wb)$\_$14-digits$\_$right.png}}"
    \\ 24bit rectified right RGB image
    \\
    \item Disparity, depth, and confidence map (1920$\times$1080 or 1280$\times$720)
    \\ ``{\bf{out$\_$zed(wb)$\_$14-digits$\_$disp.png}}"
    \\ 8bit left disparity map
    \vspace{0.1cm}
    \\ ``{\bf{out$\_$zed(wb)$\_$14-digits$\_$depth.png}}"
    \\ 16bit left depth map
    \vspace{0.1cm}
    \\ ``{\bf{out$\_$zed(wb)$\_$14-digits$\_$conf.png}}"
    \\ 8bit confidence map
    \\
    \item Calibration parameters
    \\ ``{\bf{calib$\_$zed(wb)$\_$2-digits$\_$6-digits.mat}}"
    \\ intrinsic and extrinsic camera parameters
\end{itemize}

Similar to the indoor data, each RGB-D image pair is identified by 14 digits.
A total of four ZED cameras and single built-in camera were used.
A file name includes an additional abbreviation, indicating the ZED stereo or built-in stereo cameras (for example, out$\_$zed$\_$03$\_$160524$\_$000001).
The camera parameters vary according to the date (6-digits).

\subsection{Outdoor data toolbox}
\begin{itemize}
    \item ``{\bf{compute$\_$conf.m}}" - Computes the confidence map ~\cite{Kim2018}.
\end{itemize} 
We obtain the confidence map of 400,000 stereo pairs using ~\cite{Park2015}\footnote{http://cvl.gist.ac.kr/project.} and the rest using~\cite{Kim2018}.
Note that~\cite{Kim2018} uses as inputs the cost volume and its associated disparity map, which were computed using~\cite{Zbontar2015}\footnote{https://github.com/jzbontar/mc-cnn.}.

\section{Concluding Remarks}

As future work, we will demonstrate that our dataset can indeed be used to successfully train large DNN for a wide range of computer vision and image processing applications.
We anticipate that the scale and quality of our dataset could help bridge the gap between indoor and outdoor simulations, especially for depth-related tasks, e.g., monocular depth estimation, RGB-D feature learning, outdoor scene understanding, and dehazing.

	\bibliographystyle{IEEEtran}			
	\bibliography{egbib}

\end{document}